\documentclass[11pt,a4paper]{article}
\usepackage[hyperref]{acl2018}
\usepackage{times}
\usepackage{latexsym}
\usepackage{amsmath, amsthm, latexsym, amssymb, bm, graphicx, bold-extra, mathrsfs, frcursive, algorithm}
\usepackage[noend]{algpseudocode}

\graphicspath{ {Figure/} }
\usepackage{subfigure}
\usepackage{color}
\DeclareMathOperator*{\argmin}{\arg\!\min}
\DeclareMathOperator*{\argmax}{\arg\!\max}

\usepackage{url}

\aclfinalcopy % Uncomment this line for the final submission
\title{Document Similarity for Texts of Varying Lengths via Hidden Topics}

\author{Hongyu Gong\textsuperscript{*}\quad Tarek Sakakini\textsuperscript{*} \quad Suma Bhat\textsuperscript{*} \quad Jinjun Xiong \textsuperscript{$\dagger$} \\
\textsuperscript{*}University of Illinois at Urbana-Champaign, USA \\ \textsuperscript{$\dagger$}T. J. Watson Research Center, IBM \\
\textsuperscript{*}{\{hgong6, sakakini, spbhat2\}@illinois.edu}\quad  \textsuperscript{$\dagger$}jinjun@us.ibm.com}

\date{}

\begin{document}

\maketitle

\begin{abstract}
Measuring similarity between texts is an important task for several applications. Available approaches to measure document similarity are inadequate for  document pairs that have non-comparable lengths, such as a long document and its summary. This is because of the lexical, contextual  and the abstraction gaps between a long document of rich details and its concise summary of abstract information. In this paper, we present a document matching approach to bridge this gap, by comparing the texts in a common space of hidden topics.  We evaluate the matching algorithm on two matching tasks  and find that it consistently and widely outperforms strong baselines. We also highlight the benefits of incorporating domain knowledge to text matching.
\end{abstract}

\section{Introduction}
\label{sec:introduction}

Measuring the similarity between documents is of key importance in several natural processing applications, including information retrieval \cite{salton1988term}, book recommendation \cite{gopalan2014content}, news categorization \cite{ontrup2002hyperbolic}  and essay scoring \cite{landauer2003automatic}. A range of document similarity approaches have been proposed and effectively used in recent applications, including \cite{lai2015recurrent,bordes2015large}. Central to these tasks  is the assumption that the documents being compared are of comparable lengths. %Recent media applications such as text summarization, web search \cite{mohan2011web},  processing of microblog posts, and question answering have led to new document similarity methods such as \cite{yan2013biterm,sridhar2015unsupervised}. This is because typical document sizes in these applications are  only a few hundred characters rendering   similarity measures used for comparing long documents inadequate when directly used in these tasks. Central to the tasks discussed above is the assumption that the documents being compared are of comparable lengths. 

\begin{table}[htbp!]
\centering
\small
\caption{A Sample Concept-Project Matching}
\label{tab:example}
\begin{tabular}{l}
\hline
\textbf{Concept} \\
Heredity: Inheritance and Variation of Traits \vspace{1mm}\\
\begin{tabular}[c]{@{}l@{}}All cells contain \underline{genetic} information in the form of \underline{DNA}  \\ molecules. \underline{Gene}s are regions in the DNA that contain \\the instructions that code for  the formation of \underline{proteins}.\end{tabular} \vspace{1mm} \\ \hline
\textbf{Project} \\
Pedigree Analysis: A Family Tree of Traits \vspace{1mm}\\
\begin{tabular}[c]{@{}l@{}}
Introduction: Do you have the same hair color or eye color\\ as your  mother? When we look at members of a family it\\ is easy  to see that some physical characteristics or traits\\ are shared. To start this project, you should draw a pedigree \\ showing the different members of your family. Ideally you\\ should include multiple people from at least three  \\generations.\\
%\vdots
\vspace{1mm}\\
Materials and Equipment: Paper, Pen, 
Lab notebook\\
Procedure: Before starting this science project, you should\\
%\vdots
\end{tabular}
\\ 
\hline
\end{tabular}
\end{table}

Advances in natural language understanding techniques, such as text summarization and recommendation,  have generated new requirements for comparing documents. For instance, summarization techniques (extractive and abstractive) are capable of automatically generating textual summaries by converting a long document of several hundred words into a condensed text of only a few words while preserving  the core meaning of the original text \cite{kedzie2016extractive}. %This has been used in a range of downstream applications such as generating summaries of books \cite{bamman2013new}, report generation and paper summaries \cite{jaidka2016overview}. 
Conceivably, a related aspect of  summarization is the task of bidirectional matching of a   summary and a document or a set of  documents, which is the focus of this study. The document similarity  considered in this paper is between texts that have significant differences not only in length, but also in the abstraction level  (such as a definition of an abstract concept versus a detailed instance of that abstract concept). % Such a  matching step would enable grouping of documents not only by how related they are to each other but also by how related they are to the given summary. This in turn enables recommending a new document to match the existing set of documents while also being related to the summary. Ideally, we would expect  both extractive and abstractive summaries to be matched with documents. However, considering that an abstractive summary retains the content and meaning in a more flexible and abstract manner compared to using portions of original text as done in an extractive summary, we can expect abstractive summaries to be more amenable to the matching task. This study focuses on matching of a summary with a document and vice-versa. Dissimilar texts considered in this paper are texts that have significant differences not only in the length of the texts (such as a summary versus a full article), but also in the abstraction level of the texts (such as a definition of an abstract concept versus a detailed instance of that abstract concept). Moreover, the longer texts typically exhibit more verbose words with many noisy textual information that may further obscure the underlying  true connections between the two texts.

As an illustration, consider the task of matching a {\bf Concept} with a {\bf Project} as shown in Table~\ref{tab:example}. Here a {\bf Concept} is  a grade-level science curriculum item and represents the summary.   A {\bf Project}, listed in a collection of science projects, represents the document. Typically, projects are long texts, including an introduction,  materials and procedures, whereas science concepts are significantly shorter in comparison having a title and a concise and abstract description. The concepts and projects are described in detail in Section~\ref{sec:cpmatching}. The matching task is to automatically suggest a hands-on project for a given concept in the curriculum,  such that the project  can help reinforce a learner's basic understanding of the concept. Conversely, given a science project, one may need to identify the concept it covers by matching it to a listed concept in the curriculum. This would be conceivable in the context of an intelligent tutoring system.

Challenges to the matching task mentioned above include: 1) A mismatch in the relative lengths of the documents being compared -- a long piece of text (henceforth termed \textit{document}) and a short piece of text (termed  \textit{summary}) -- gives rise to the vocabulary mismatch problem, where the document and the summary do not share a majority of key terms. 2)  The context mismatch problem arising because a document provides a reasonable amount of text to infer the contextual meaning of a key term, but a summary only provides a limited context, which may or may not involve the same terms considered in the document.  %2) The summary may be an abstraction of the content of the document, as in the case where a summary (the {\bf Concept} in the example considered) captures the key concepts pertaining to a document (the {\bf Project}). 
These challenges render existing approaches to  comparing documents--for instance, those that rely on document representations (e.g., Doc2Vec \cite{le2014distributed})--inadequate, because the predominance of non-topic words in the document introduces noise to its representation while the summary is relatively noise-free, rendering Doc2Vec inadequate for comparing them.   

Our approach to the matching problem is to allow a multi-view generalization of the document. Here, multiple hidden topic vectors are used to establish a common ground to capture as much information of the document and the summary as possible and  eventually  to score the  relevance of the pair. We empirically  validate  our approach on two tasks -- that of project-concept matching in grade-level science and that of scientific paper-summary matching--using both custom-made and publicly available datasets. The main contributions of this paper are:
\begin{itemize}
\item We propose an embedding-based hidden topic model to extract topics and measure their importance in long documents.
\item We present a novel geometric approach to compare documents with differing modality (a long document to a short summary) and  validate its performance relative to strong baselines. 
\item We explore the use of domain-specific word embeddings for the matching task and show the explicit benefit of incorporating  domain knowledge in the algorithm.
\item We make available the first dataset\footnote{Our code and data are available at: \url{https://github.com/HongyuGong/Document-Similarity-via-Hidden-Topics.git}} on project-concept matching in the science domain to help further research in this area.
\end{itemize}

\section{Related Works}
\label{sec:relatedWork}

% domain knowledge
% ??? \cite{cheung2013probabilistic} pointed out that in information extrac- tion, a list of slots in the target domain is given to the system, and in natural language generation, content models are trained to learn the content structure of texts in the target domain for infor- mation structuring and automatic summarization

%\textbf{Domain knowledge.} Languages vary a lot from domain to domain, and texts from different domains have different vocabularies and semantic styles. Domain-specific information could play a role of disambiguation and denoising in language understanding. It has been shown to be critical in many NLP applications. Domain modeling is a key part in  in information extraction and multi-document summarization \cite{cheung2013probabilistic}. It is pointed out that spoken language understanding \cite{chen2015matrix} and aspect extraction \cite{chen2013exploiting} are domain-specific. 

\begin{figure*}[htbp!]
\centering
\begin{minipage}{0.48\textwidth}
\centerline{\includegraphics[width=0.8\linewidth]{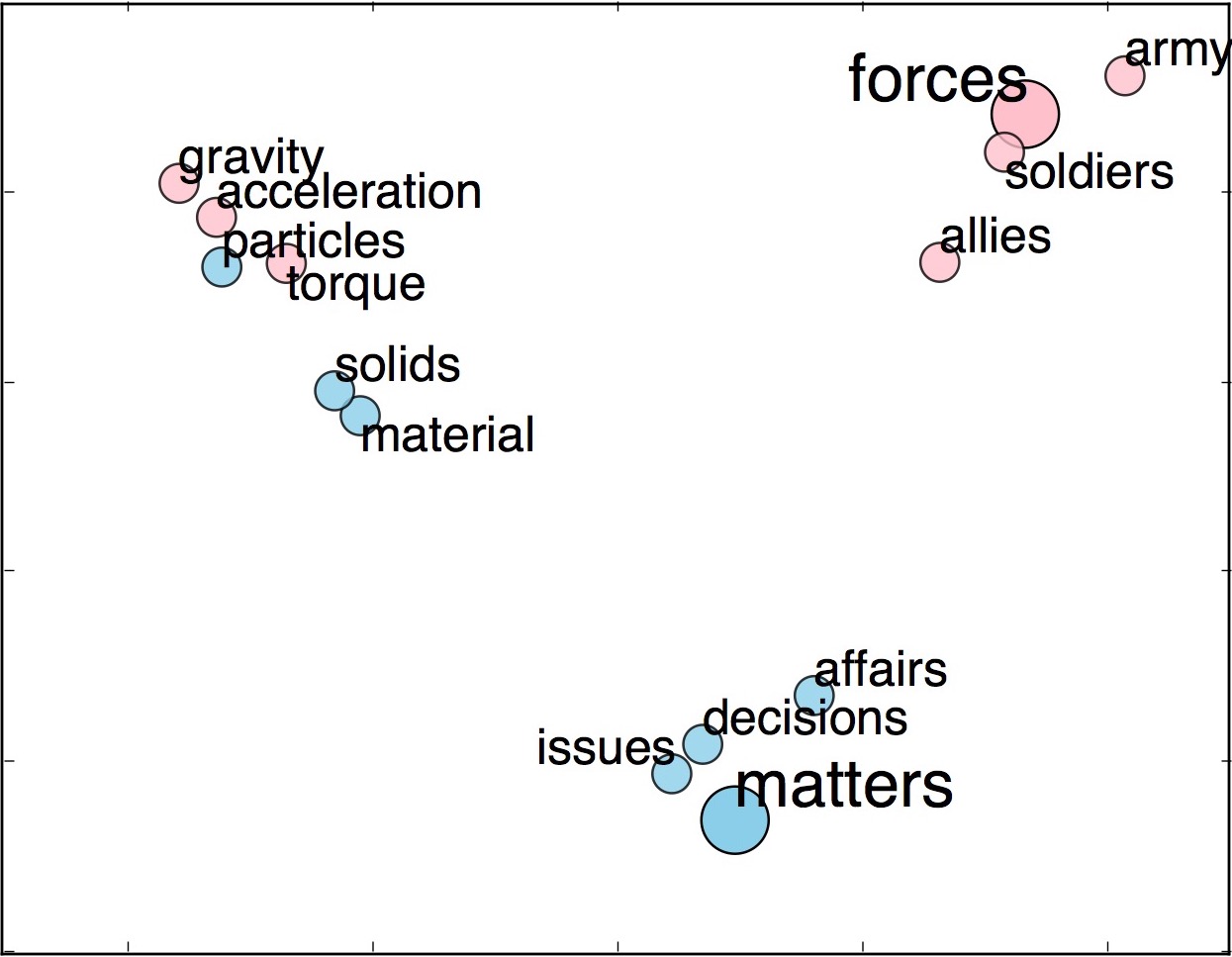}}
\centerline{\small{(a) word geometry of general embedding}}
\label{fig:globaCluster}
\end{minipage}
\begin{minipage}[c]{0.48\textwidth}
\centerline{\includegraphics[width=0.8\linewidth]{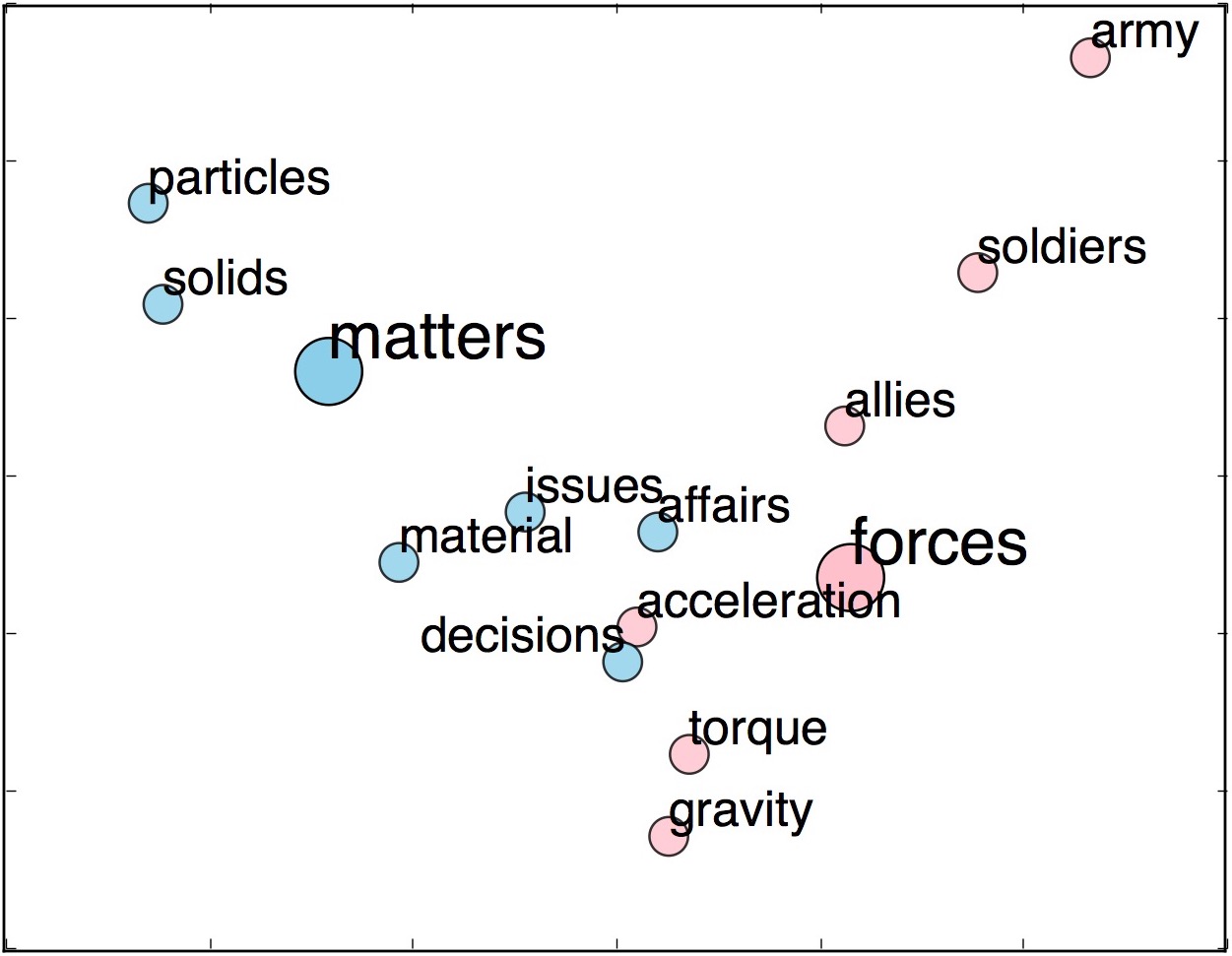}}
\centerline{\small{(b) word geometry of science domain embeddings}}
\label{fig:domainCluster}
\end{minipage}
\caption{Two key words ``forces'' and ``matters'' are shown in red and blue respectively. Red words represent different senses of ``forces'', and blue words carry senses of ``matters''. ``forces'' mainly refers to ``army'' and ``matters'' refers to ``issues'' in general embedding of (a), whereas ``forces'' shows its sense of ``gravity'' and ``matters'' shows the sense of ``solids'' in science-domain embedding of (b)}
\label{fig:globalDomainSense}
\end{figure*}
%\vspace{-8mm}

%It was found that statistical models without domain knowledge led to uninterpretable aspects in sentiment analysis. 
%Summarization is also a domain-dependent task, and in-domain documents made it possible to reconstruct model summaries from source text \cite{cheung2013towards}. In this paper, we will show that science domain has a great influence on lexical semantics, and further incorporate domain knowledge into the concept-project matching system. 

% relatedness of two documents
%\noindent\textbf{Document similarity} is a straightforward way to quantify the relevance between projects and concepts. Here we will discuss recent works on document similarity. One group of methods are embedding based, i.e., representing each document as a vector, and cosine distance is a commonly used similarity metric. Given vector representations $v_{1}$ and $v_{2}$ for documents $d1$ and $d2$, their similarity is $\text{sim}(d1,d2)=\frac{v_{1}^{T}v_{2}}{\lVert v_{1}\rVert \lVert v_{2}\rVert}$.

\textbf{Document similarity} approaches quantify the degree of relatedness between two pieces of texts of comparable lengths and thus enable matching between documents.  Traditionally,  statistical approaches (e.g., \cite{metzler2007similarity}) and vector-space-based methods (including the robust  Latent Semantic Analysis (LSA) \cite{dumais2004latent}) have been used.  More recently, neural network-based methods have been used for document representation and comparison. These methods  include average word embeddings \cite{mikolov2013distributed}, Doc2Vec \cite{le2014distributed}, Skip-Thought vectors \cite{kiros2015skip}, recursive neural network-based methods \cite{socher2014grounded}, LSTM architectures \cite{tai2015improved}, and convolutional neural networks \cite{blunsom2014convolutional}. %We avoid using document representation due to its computational complexity and due to the fundamental difference in word choice and lengths between documents concepts. 
%Given vector representations $v_{1}$ and $v_{2}$ for documents $d1$ and $d2$, their similarity is $\text{sim}(d1,d2)=\frac{v_{1}^{T}v_{2}}{\lVert v_{1}\rVert \lVert v_{2}\rVert}$.

%Document representation is a widely studied field due to being a fundamental stage in many NLP tasks like text classification and sentiment analysis. 

%The distance between the key word and its neighbor reflects how related their embedding is to the corresponding sense.

Considering works that avoid using an explicit document representation for comparing documents, the state-of-the-art method is word mover's distance (WMD), which relies on pre-trained word embeddings \cite{kusner2015word}. Given these embeddings, WMD defines the distance between two documents as the best transport cost of moving all the words from one document to another within the space of word embeddings. % Let $c(i,j)$ be the Euclidean distance between two word vectors $v_{i}$ and $v_{j}$. Let $d_{i}$ and $d_{j}'$ be the normalized frequencies of word $i$ and $j$ in documents $d$ and $d'$ respectively. Denote $T_{ij}$ as the percentage of word $i$ in document $d$ that needs to be replaced with word $j$ to reach document $d'$. The Word Mover's distance is the minimum cost to move words in one document to words in the other document, formulated as:
%\begin{comment}
%\begin{align}
%\label{eq:1}
%\min\limits_{T\ge 0} &\sum\limits_{i,j}T_{i,j}c(i,j) \\
%\nonumber
%\text{subject to:} &\sum\limits_{j}T_{ij}=d_{i}, \forall i \\
%\nonumber
%&\sum\limits_{i}T_{ij} = d_{j}', \forall j
%\end{align}
%\end{comment}
The advantages of WMD  are that it is free of hyper-parameters and achieves high retrieval accuracy on document classification tasks with documents of comparable lengths. However, it is computationally expensive for long documents as mentioned in \cite{kusner2015word}.

Clearly, what is lacking in prior literature is a study of  matching approaches for documents with widely different sizes. It is this gap in  the literature that we expect to fill by way of this study. %This is why we also avoid using WMD as is.

\noindent\textbf{Latent Variable Models}. Latent variable models including count-based and probabilistic models have been studied in many previous works. Count-based models such as Latent Semantic Indexing (LSI) compare two documents based on their combined vocabulary \cite{deerwester1990indexing}.
When documents have highly mismatched vocabularies such as those that we study, relevant documents might be classified as irrelevant. Our model is built upon word-embeddings which is more robust to such a vocabulary mismatch.

Probabilistic models such as Latent Dirichlet Analysis (LDA) define topics as distributions over words \cite{blei2003latent}. In our model, topics are low-dimensional real-valued vectors (more details in Section~\ref{hidden_vectors}).

\begin{figure*}[ht!]
\centering
\includegraphics[width=0.7\textwidth]{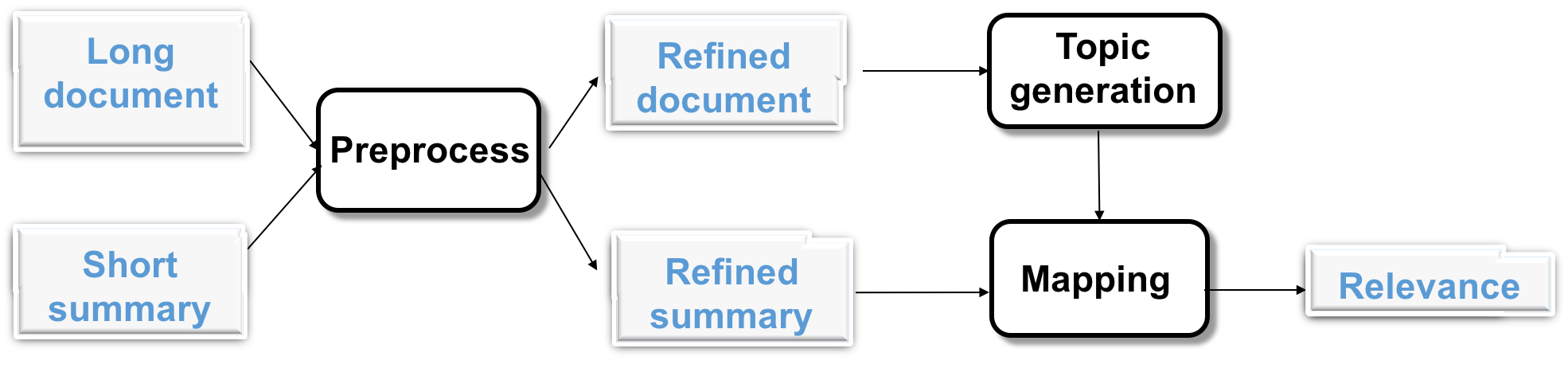}
\caption{The system for document-summary matching}
\label{fig:system}
\end{figure*}

\section{Domain Knowledge}

Domain information pertaining to specific  areas of knowledge is made available in texts by the use of words with domain-specific meanings or senses.  
%This results in the same word taking on a multiplicity of senses depending on the domain in which it is used. This also means that domain-specific information plays a major role in resolving the ambiguity inherent in words. 
Consequently, domain knowledge has been shown to be critical in many NLP applications such as information extraction and multi-document summarization \cite{cheung2013probabilistic},  spoken language understanding \cite{chen2015matrix}, aspect extraction \cite{chen2013exploiting} and summarization \cite{cheung2013towards}. 

%In line with previous findings, and more specifically in the context of this work, we will show how a domain-specific corpus allows for a better modeling of lexical semantics in comparison to a general corpus, while matching between a document and a summary. 

As will be described later, our distance metric for comparing a document and a summary relies on word embeddings. 
We show in this work, that embeddings trained on a science-domain corpus lead to better performance than embeddings on the general corpus (WikiCorpus). 
Towards this, we extract a science-domain sub-corpus from the WikiCorpus, and the corpus extraction will be detailed in Section \ref{sec:exp}.

To motivate the domain-specific behavior of polysemous words, we  will qualitatively explore how domain-specific embeddings differ from the general embeddings on two polysemous science terms: \emph{forces} and \emph{matters}. Considering the fact that the meaning of a word is dictated by its neighbors, for each set of word embeddings, we plot the neighbors of these two terms in Figure~\ref{fig:globalDomainSense} on to 2 dimensions using Locally Linear Embedding (LLE), which preserves word distances \cite{roweis2000nonlinear}. We then analyze the sense of the focus terms--here, \emph{forces} and \emph{matters}.

From Figure~\ref{fig:globalDomainSense}(a), we see that for the word \emph{forces}, its general embedding is close to \emph{army}, \emph{soldiers}, \emph{allies}  indicating that it is related with violence and power in a general domain.  Shifting our attention to  Figure~\ref{fig:globalDomainSense}(b), we see that  for the same term, its science embedding is closer to \emph{torque}, \emph{gravity}, \emph{acceleration}  implying that its science sense is more about physical interactions. Likewise,  for the word \emph{matters}, its  general embedding is surrounded by \emph{affairs} and \emph{issues}, whereas, its science embedding is closer to \emph{particles} and \emph{material}, prompting that it represents substances. Thus, we conclude that  domain specific embeddings (here, science), is  capable of incorporating domain knowledge into word representations. We use this observation in our document-summary matching system to which we turn next.

\section{Model}
\label{sec:model}

Our model that performs the matching between document and summary  is depicted in Figure~\ref{fig:system}. It is composed of three modules that perform preprocessing, document topic generation, and relevance measurement between a document and a summary. Each of these modules is discussed below.

\subsection{Preprocessing}

%Preprocessing is to tokenize the texts, as well as to remove stop words such as articles, pronouns, numbers, conjunction and prepositions from the text.  After preprocessing, we have refined texts for each project and concept, which contain only content words.

The preprocessing module tokenizes  texts and removes stop words and prepositions. This step allows our system to focus on the content words without impacting the meaning of original texts. 

% remove tf-idf part
% % \subsection{TF-IDF Model}
% % TF-IDF, short for term frequency-inverse document frequency, measures the importance of a word in a document \cite{salton1988term}. Suppose that a word or a term is represented as $t$, a document as $d$ and a collection of documents as corpus $c$.
% % Term frequency, denoted as $\text{tf}(t,d)$, is the frequency of term $t$ in document $d$. 
% % The inverse document frequency is defined as: 
% % $\text{idf}(t)=1+\log\frac{1+|c|}{1+n(d',t)}$, where $|c|$ is the total number of documents in $c$, and $n(d',t)$ is the number of documents containing $t$. When a word occurs many times in a document, but not frequently used in other documents, it is likely that this word is informative. We apply the tf-idf model to the corpus of science projects, and keep $500$ words with the highest tf-idf score in each project. TF-IDF module is used to denoise long project descriptions.

\subsection{Topic Generation from Documents}
\label{hidden_vectors}
We assume that a document (a long text) is a structured collection of words, with  the `structure' brought about by the composition of topics. In some sense, this `structure' is represented as a set of hidden topics. Thus,  we assume that a document   is generated from certain hidden ``topics'', analogous to the modeling assumption in  LDA. However, unlike in LDA,  the ``topics'' here are neither specific words nor the distribution over words, but are  are essentially a set of \emph{vectors}. In turn, this means that words (represented as vectors) constituting the document  structure can be generated from the hidden topic vectors.

Introducing some notation, the word vectors in a document are $\{{\bf w}_{1}, \ldots, {\bf w}_{n}\}$, and the hidden topic vectors of the document are $\{{\bf h}_{1},\ldots, {\bf h}_{K}\}$, where ${\bf w}_{i}, {\bf h}_{k}\in\mathbb{R}^{d}$, $d=300$ in our experiments.

Linear operations using word embeddings have been empirically shown to approximate   their compositional properties (e.g. the embedding of a phrase is nearly the sum of the embeddings of its component words) \cite{mikolov2013distributed}. This motivates the linear reconstruction of the words from the document's hidden topics while minimizing the reconstruction error. 
We stack the $K$ topic vectors as a topic matrix ${\bf H}=[{\bf h}_{1},\ldots,{\bf h}_{K}] (K<d)$. We define the reconstructed word vector $\tilde{{\bf w}}_{i}$ for the word ${\bf w}_{i}$ as the optimal linear approximation given by topic vectors: $\tilde{{\bf w}}_{i} = {\bf H}\tilde{{\bm \alpha}}_{i}$, where \\
\begin{align}
\label{eq:alpha}
\tilde{{\bm \alpha}}_{i}=\argmin\limits_{{\bm \alpha}_{i}\in\mathbb{R}^{K}}\lVert {\bf w}_{i}-{\bf H}{\bm\alpha}_{i}\rVert_{2}^{2}.
\end{align}

The reconstruction error $E$ for the whole document  is the sum of each word's reconstruction error and is given by:
$E = \sum\limits_{i=1}^{n}\lVert{\bf w}_{i}-\tilde{{\bf w}}_{i}\rVert_{2}^{2}$. 
%Without loss of generality, we assume that these topic vectors are orthonormal, i.e., ${\bf h}_{i}^{T}{\bf h}_{j}=1_{(i==j)}$, since the orthonormal constraints do not change the linear space formed by topic vectors.
  This being a function of the topic vectors, our goal is to find the optimal ${\bf H}^{*}$ so as to minimize the error $E$:
\begin{align}
\nonumber
{\bf H}^{*} &=\argmin\limits_{{\bf H}\in\mathbb{R}^{d\times K}} E({\bf H}) \\
\label{eq:2}
&= \argmin\limits_{{\bf H}\in\mathbb{R}^{d\times K}}\sum\limits_{i=1}^{n}\min\limits_{{\bm \alpha}_{i}}\lVert {\bf w}_{i}-{\bf H}{\bm\alpha}_{i}\rVert_{2}^{2},
\end{align}
where $\lVert\cdot\rVert$ is the Frobenius norm of a matrix.

Without loss of generality, we require the topic vectors  $\{{\bf h}_{i}\}_{i=1}^{K}$ to be orthonormal, i.e., ${\bf h}_{i}^{T}{\bf h}_{j}=1_{(i=j)}$. As we can see, the optimization problem (\ref{eq:2}) describes an optimal linear space spanned by the topic vectors, so the norm and the linear dependency of the vectors do not matter.
With the orthonormal constraints, we simplify the form of the reconstructed vector $\tilde{{\bf w}}_{i}$ as: 
\begin{align}
\label{eq:reconstruct}
\tilde{{\bf w}}_{i}={\bf H}{\bf H}^{T}{\bf w}_{i}. 
\end{align}
We stack word vectors in the document as a matrix ${\bf W}=[{\bf w}_{1}, \ldots, {\bf w}_{n}]$. The equivalent formulation to problem (\ref{eq:2}) is:
\begin{align}
\nonumber
&\min\limits_{{\bf H}}\quad \lVert {\bf W} - {\bf H}{\bf H}^{T}{\bf W}\rVert^{2}_{2} \\
\label{eq:matrix}
&\text{s.t.}\quad {\bf H}^{T}{\bf H}={\bf I},
\end{align}
where ${\bf I}$ is an identity matrix.

The problem can be solved by Singular Value Decomposition (SVD), using which, the matrix ${\bf W}$ can be decomposed as ${\bf W}={\bf U}{\bm\Sigma}{\bf V}^{T}$, where ${\bf U}^{T}{\bf U}={\bf I}$,${\bf V}^{T}{\bf V}={\bf I}$, and ${\bm \Sigma}$ is a diagonal matrix  where the diagonal elements are arranged in a decreasing order of absolute values. We show in the supplementary material that the first $K$ vectors in the matrix $\bf U$ are exactly the solution to ${\bf H}^{*}=[{\bf h}_{1}^{*}, \ldots, {\bf h}_{K}^{*}]$. %A detailed proof is included in the supplementary material. 

We find optimal topic vectors ${\bf H}^{*}=[{\bf h}_{1}^{*}, \ldots, {\bf h}_{K}^{*}]$ by solving problem (\ref{eq:matrix}). We note that these topic vectors are not equally important, and we say that one topic is more important than another if it can reconstruct words with smaller error. Define $E_{k}$ as the reconstruction error when we only use topic vector ${\bf h}^{*}_{k}$ to reconstruct the document:
\begin{align}
\label{eq:topic_err}
E_{k} = \lVert {\bf W}-{\bf h}_{k}^{*}{{\bf h}_{k}^{*}}^{T}{\bf W}\rVert_{2}^{2}.
\end{align}
Now  define $i_{k}$ as the \emph{importance} of topic ${\bf h}_{k}^{*}$, which measures the topic's ability to reconstruct the words in a document:
\begin{align}
\label{eq:important}
i_{k}=\lVert {{\bf h}_{k}^{*}}^{T}{\bf W}\rVert_{2}^{2}
%i_{k}=\lVert {\bf h}_{k}^{*}{{\bf h}_{k}^{*}}^{T}{\bf W}\rVert_{2}^{2}.
\end{align}

We show in the supplementary material that the higher the importance $i_{k}$ is, the smaller the reconstruction error $E_{k}$ is. Now we normalize $i_{k}$ as $\bar{i}_{k}$ so that the importance does not scale with the norm of the word matrix $W$, and so that the importances of the $K$ topics sum to $1$. Thus, 
\begin{align}
\label{eq:importance}
\vspace{-4pt}
\bar{i}_{k} = i_{k}/(\sum\limits_{j=1}^{K}i_{j}).
\end{align}
{The number of topics $K$ is a hyperparameter in our model. A small $K$ may not cover key ideas of the document, whereas a large $K$ may keep trivial and noisy information. Empirically we find that $K=15$ captures most important information from the document.

\subsection{Topic Mapping to Summaries}
We have extracted $K$ topic vectors $\{{\bf h}_{k}^{*}\}_{k=1}^{K}$ from the document matrix ${\bf W}$, whose importance is reflected by $\{\bar{i}_{k}\}_{k=1}^{K}$. In this module, we measure the relevance of a document-summary pair. Towards this, a summary that matches the document should also be closely related with the ``topics'' of that document. 
Suppose the vectors of the words in a summary are stacked as a $d\times m$ matrix ${\bf S}=[{\bf s}_{1},\ldots, {\bf s}_{m}]$, where ${\bf s}_{j}$ is the vector of the j-th word in a summary. Similar to the reconstruction of the document, the summary can also be reconstructed from the documents' topic vectors as shown in Eq. (\ref{eq:reconstruct}). 
% The reconstruction of word ${\bf s}_{j}$ by $K$ topics are 
% \begin{align}
% \label{eq:word}
% \tilde{{\bf s}}_{j}=H^{*}{H^{*}}^{T}{\bf s_{j}}.
% \end{align}
Let $\tilde{{\bf s}}_{j}^{k}$ be the reconstruction of the summary word ${\bf s}_{j}$ given by one topic ${\bf h}_{k}^{*}$:
$\tilde{{\bf s}}_{j}^{k}={\bf h}_{k}^{*}{{\bf h}_{k}^{*}}^{T}{\bf s}_{j}.$

Let $r({\bf h}_{k}^{*}, {\bf s}_{j})$ be the relevance between a topic vector ${\bf h}_{k}^{*}$ and summary word ${\bf s}_{j}$. It is defined as the cosine similarity between $\tilde{{\bf s}}_{j}^{k}$ and ${\bf s}_{j}$:
\begin{align}
\label{eq:rel_word}
r({\bf h}_{k}^{*}, {\bf s}_{j}) = {{\bf s}_{j}}^{T}\tilde{s}_{j}^{k}/(\lVert{\bf s}_{j}\rVert_{2}\cdot\lVert\tilde{{\bf s}}_{j}^{k}\rVert_{2}).
\end{align}
Furthermore, let $r({\bf h}_{k}^{*}, {\bf S})$ be the relevance between a topic vector and the summary, defined to be the average similarity between the topic vector and the summary words:
\begin{align}
\label{eq:rel_summ}
r({\bf h}_{k}^{*}, {\bf S}) = \frac{1}{m}\sum\limits_{j=1}^{m}r({\bf h}_{k}^{*}, {\bf s}_{j}).
\end{align}
The relevance between a topic vector and a summary is a real value between 0 and 1.

As we have shown, the topics extracted from a document are not equally important. Naturally, a summary relevant to more important topics is more likely to better match  the document. Therefore, we define $r({\bf W},{\bf S})$ as the relevance between the document ${\bf W}$ and the summary ${\bf S}$, and $r({\bf W},{\bf S})$ is the sum of topic-summary relevance weighted by the importance of the topic:
\begin{align}
\label{eq:rel_doc_summ}
r({\bf W},{\bf S}) = \sum\limits_{k=1}^{K}\bar{i}_{k}\cdot r({\bf h}^{*}_{k}, {\bf S}),
\end{align}
where $\bar{i}_{k}$ is the importance of topic ${\bf h}^{*}_{k}$ as defined in (\ref{eq:importance}). The higher $r({\bf W},{\bf S})$ is, the better the summary matches the document.

We provide a visual representation of the documents as shown in  Figure~\ref{fig:weather_heredity} to illustrate the notion of hidden topics. The two documents are from science projects: a genetics project, \emph{Pedigree Analysis: A Family Tree of Traits} \cite{pedigree}, and a weather project, \emph{How Do the Seasons Change in Each Hemisphere} \cite{weather}. We project all embeddings to a three-dimensional space for ease of visualization.

\begin{figure}[htbp!]
\centering
\includegraphics[width=0.38\textwidth]{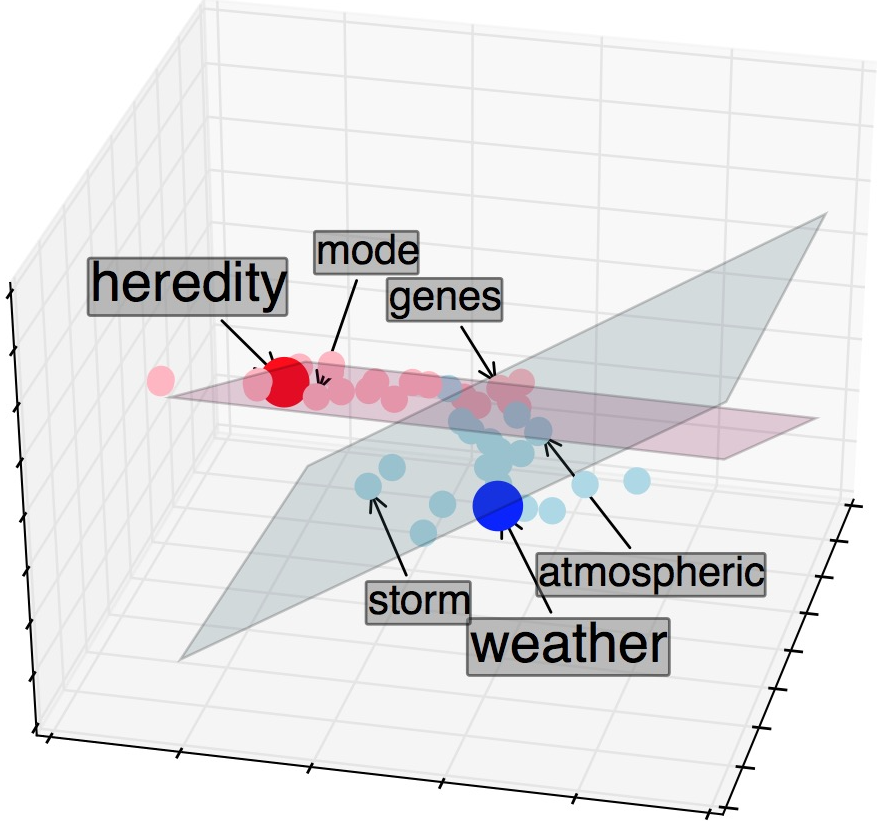}
\caption{Words \emph{mode} and \emph{genes} from the document on genetics and words \emph{storm} and \emph{atmospheric} from document on weather are represented by pink and blue points respectively. Linear space of hidden topics in genetics form the pink plane, where summary word \emph{heredity} (the red point) roughly lies. Topic vectors of the document on weather form the blue plane, and the summary word \emph{weather} (the darkblue point) lies almost on the same plane.}
\label{fig:weather_heredity}
\vspace{-4mm}
\end{figure}
%\vspace{-4mm}

As seen in Figure~\ref{fig:weather_heredity}, the hidden topics reconstruct the words in their respective documents to the extent  possible. This means that the words of a document lie roughly on the plane formed by their corresponding topic vectors. We also notice that the summary words (\emph{heredity} and \emph{weather} respectively for the two projects under consideration) lie very close to the  plane formed by the hidden topics of the relevant project while remaining away from the plane of the irrelevant project. This shows that the words in the summary (and hence the summary itself) can also be reconstructed from the hidden topics of documents that match the summary (and are hence `relevant' to the summary). Figure~\ref{fig:weather_heredity} visually explains the geometric relations between the summaries, the hidden topics and the documents. It also validates the representation power of the extracted hidden topic vectors. 
%because of their ability to generate abstractive summaries.

\section{Experiments}
\label{sec:exp}
In this section, we evaluate our document-summary matching approach  on two specific applications where texts of different sizes are compared. One application is that of concept-project matching useful in science education and the other is that of summary-research paper matching.

\textbf{Word Embeddings}. Two sets of 300-dimension word embeddings were used in our experiments. They were trained by the Continuous Bag-of-Words (CBOW) model in word2vec \cite{mikolov2013distributed} but on different corpora. One training corpus is the full English WikiCorpus of size $9$ GB \cite{polyglot:2013:ACL-CoNLL}. The second consists of science articles extracted from the WikiCorpus. To extract these science articles, we manually selected the science categories in Wikipedia and considered all subcategories within a depth of  3 from these manually selected root categories. We then extracted all articles in the aforementioned science categories resulting in a science corpus of size $2.4$ GB. The  word vectors used for documents and summaries are both from the pretrained word2vec embeddings.

\noindent\textbf{Baselines} We include two state-of-the-art methods of measuring document similarity for comparison using their implementations  available in gensim \cite{rehurek_lrec}. \\
(1) \textbf{Word movers' distance (WMD)} \cite{kusner2015word}. WMD quantifies the distance between a pair of documents based on word embeddings as introduced previously (c.f. Related Work). We take the negative of their distance as a measure of document similarity (here between a document and a summary). \\
%The smaller the distance is, the more similar documents are.  \\
(2) \textbf{Doc2Vec} \cite{le2014distributed}.  Document representations have been trained with neural networks. We used two versions of doc2vec: one trained on the full English Wikicorpus and a second trained on the science corpus, same as the corpora used for word embedding training.
We used the cosine similarity between two text vectors to measure their relevance.

For a given document-summary pair, we compare the scores obtained using the above two methods with that obtained using our method. 

\subsection{Concept-Project matching}
\label{sec:cpmatching}

%Experiential learning (learning by doing) is essential for learners' comprehension of complex concepts.
Science projects are valuable resources for learners to instigate knowledge creation via experimentation and observation. The need for matching a science concept with a science project arises when learners intending to delve deeper into certain concepts  search for  projects that match a given concept. Additionally, they may want to identify the concepts with which a set of projects are related. 

We note that in this task, science concepts are highly concise summaries of the core ideas in projects, whereas projects are detailed instructions of the experimental procedures, including an introduction, materials and a description of the procedure, as shown in Table~\ref{tab:example}. Our matching method provides a way to bridge the gap between abstract concepts and detailed projects. The format of the concepts and the projects is discussed below. 

\begin{table*}[htbp!]
\centering
\caption{Classification results for the Concept-Project Matching task. All performance differences were statistically significant at $p=0.01$.}
\label{tab:classification}
\resizebox{1.0\textwidth}{!}{
\begin{tabular}{|c|c|c|c|c|c|c|}
\hline
method & topic\_science & topic\_wiki & wmd\_science & wmd\_wiki & doc2vec\_science & doc2vec\_wiki \\ \hline
precision & ${\bf 0.758\pm 0.012}$ & $0.750\pm 0.009$ & $0.643\pm 0.070$ & $0.568\pm 0.055$ & $0.615\pm 0.055$ & $0.661\pm 0.084$ \\ \hline
recall & ${\bf 0.885\pm 0.071}$ & $0.842\pm 0.010$ & $0.735\pm 0.119$ & $0.661\pm 0.119$ & $0.843\pm 0.066$ & $0.737\pm 0.149$ \\ \hline
fscore & ${\bf 0.818\pm 0.028}$ & $0.791\pm 0.007$ & $0.679\pm 0.022$ & $0.595\pm 0.020$ & $0.695\pm 0.019$ & $0.681\pm 0.032$ \\ \hline
\end{tabular}}
\end{table*}

\noindent\textbf{Concepts}. For the purpose of this study we use the concepts available in the Next Generation Science Standards (NGSS) \cite{NGSS}.
%disciplinary core ideas (DCIs) which are key concepts in  disciplines. 
Each concept is accompanied by a short description. For example, one concept in life science is \emph{Heredity: Inheritance and Variation of Traits}. Its description is \emph{All cells contain genetic information in the form of DNA molecules. Genes are regions in the DNA that contain the instructions that code for the formation of proteins.} Typical lengths of concepts are around 50 words. \\
\textbf{Projects}. The website Science Buddies \cite{ScienceBuddies} provides a list of projects from a variety of science and engineering disciplines such as physical sciences, life sciences and social sciences. A typical project consists of an abstract, an introduction, a description of the experiment and the associated  procedures. A project typically has more than 1000 words.\\
\textbf{Dataset}. We prepared a representative dataset $537$ pairs of projects and concepts involving $53$ unique concepts from NGSS and $230$ unique projects from Science Buddies. %The typical length of concepts in our collection is around $100$ words, and that of projects is more than $1000$. 
Engineering undergraduate students annotated each pair with the decision whether it was a good match or not and received research credit. As a result, each concept-project pair received at least three annotations, and upon consolidation, we considered a concept-project pair to be a good match when a majority of the annotators agreed. Otherwise it was not considered a good match. The ratio between good matches and bad matches in the collected data was $44:56$.

\noindent\textbf{Classification Evaluation.} Annotations from students provided the ground truth labels for the classification task. We randomly split the dataset into tuning and test instances with a ratio of $1:9$. 
A threshold score was tuned on the tuning data, and concept-project pairs with  scores higher than this threshold were classified as a good matches during testing. We performed 10-fold cross validation, and report the average precision, recall, F1 score and their standard deviation in Table~\ref{tab:classification}. 

Our topic-based metric is denoted as ``topic'', and the general-domain and science-domain embeddings are denoted as ``wiki'' and ``science'' respectively. We show the performance of our method against the two baselines while varying the underlying embeddings, thus resulting in 6 different combinations. For example, ``topic\_science'' refers to our method with science embeddings. From the table (column~1) we  notice the following: 1)  Our method \textit{significantly outperforms}  the two baselines by a wide margin ($\approx$10\%) in both the general domain setting as well as the domain-specific setting. 2)  Using science domain-specific word embeddings instead of the general word embeddings results in the \textit{best performance across all algorithms}. This performance was observed despite the word embeddings being trained on a \textit{significantly smaller} corpus compared to the general domain corpus. 

Besides the classification metrics, we also evaluated the directed matching from concepts to projects with ranking metrics.

\begin{figure*}[htbp!]
\centering
\begin{minipage}{0.32\textwidth}
\centerline{\includegraphics[width=1.0\linewidth]{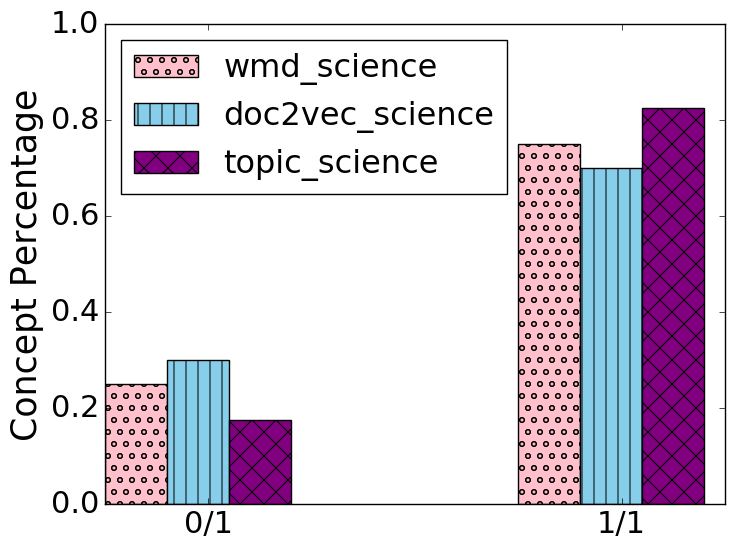}}
\centerline{\small{(a) Precision@1}}
\label{fig:rankScience1}
\end{minipage}
\begin{minipage}{0.32\textwidth}
\centerline{\includegraphics[width=1.0\linewidth]{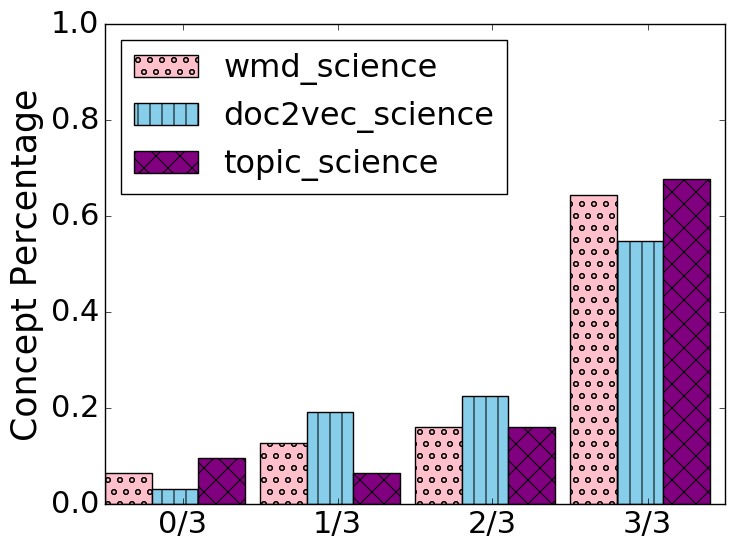}}
\centerline{\small{(b) Precision@3}}
\label{fig:rankScience3}
\end{minipage}
\begin{minipage}[c]{0.32\textwidth}
\centerline{\includegraphics[width=1.0\linewidth]{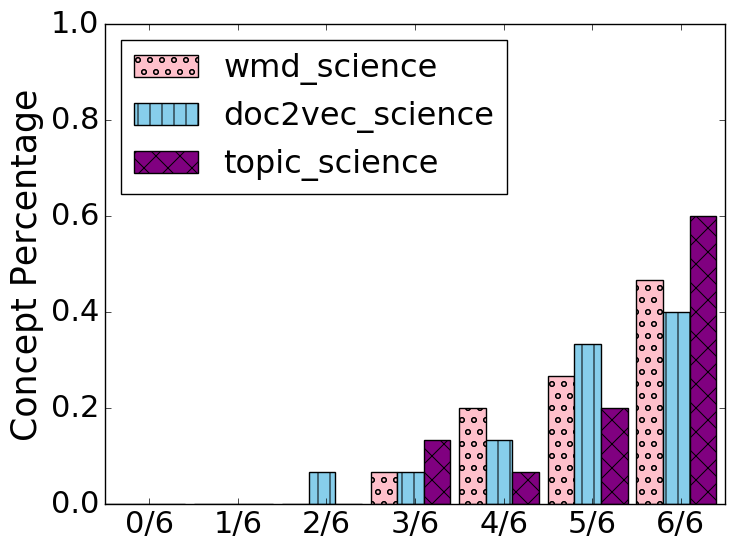}}
\centerline{\small{(c) Precision@6}}
\label{fig:ranking}
\end{minipage}
\caption{Ranking Performance of All Methods}
\label{fig:ranking}
\vspace{-5mm}
\end{figure*}

\noindent\textbf{Ranking Evaluation} Our collected dataset resulted in having  a many-to-many matching between concepts and projects. This is because the same concept was found to be a good match for multiple projects and the same project was found to match many concepts. The previously described classification task evaluated the bidirectional concept-project matching. Next we evaluated the directed matching from concepts to projects, to see how relevant these top ranking projects are to a given input concepts. Here we use precision@k \cite{radlinski2010comparing} as the evaluation metric, considering the percentage of relevant ones among top-ranking projects.

% and it is defined as:
% \begin{align*}
% \text{precision@k}=\frac{\text{\# of relevant project of ranking $<$ k}}{\text{k}}.
% \end{align*}

%For this part, we only considered the methods using science domain embeddings as they have shown superior performance in the classificaiton task. We first calculated the precision@k $(k=1,3,6)$ for concepts with at least $k$ matching projects. Then we count the percentage of concepts that fall into different value bins of precision@k. The bar graphs in Fig.~\ref{fig:ranking} show the distribution of precision@k of different methods.

For this part, we only considered the methods using science domain embeddings as they have shown superior performance in the classificaiton task. For each concept, we check the precision@k of matched projects and place it in one of k+1 bins accordingly. For example, for k=3, if only two of the three top projects are a correct match, the concept is placed in the bin corresponding to $2/3$. In Figure~ \ref{fig:ranking}, we show the percentage of concepts that fall into each bin for the three different algorithms for k=1,3,6.

We observe that recommendations using the hidden topic approach fall more in the high value bin compared to  others, performing consistently better than two strong baselines. The advantage becomes more obvious at precision@6. It is worth mentioning that wmd\_science falls behind doc2vec\_science in the classification task while it outperforms in the ranking task.

\subsection{Text Summarization}
The task of matching summaries and documents is commonly seen in real life. For example, we use an event summary ``Google's AlphaGo beats Korean player Lee Sedol in Go'' to search for relevant news, or use the summary of a scientific paper to look for  related research publications. Such matching constitutes  an ideal task to evaluate our matching method between texts of different sizes. 

\noindent\textbf{Dataset}. We use a dataset from the CL-SciSumm Shared Task \cite{jaidka2016overview}. The dataset consists of 730 ACL Computational Linguistics research papers covering 50 categories in total. Each category consists of a reference paper (RP) and around 10 citing papers (CP) that contain citations to the RP.  A human-generated summary for the RP is provided and we use the 10 CP as being relevant to the summary. %This human summary is an abstractive summarization, which gives a compressed paraphrase of the paper.
%different from extractive summary which selects key words from the source text. 
The matching task here is between the summary and all CPs in each category.\\
% \begin{figure}[t]
% \centering
% \includegraphics[width=0.48\textwidth]{summarization_ranking.png}
% \vspace{-12mm}
% \caption{Evaluation of Summary-Article matching}
% \label{fig:summarization_ranking}
% \vspace{-5mm}
% \end{figure}
\noindent\textbf{Evaluation}. For each paper, we keep all of its content except the sections of experiments and acknowledgement (these sections were omitted because often their content is often less related to the topic of the summary). The typical summary length  is about $100$ words, while a paper has more than $2000$ words. For each topic, we  rank all $730$ papers in terms of their relevance generated by our method and baselines using both sets of embeddings. For evaluation, we use the information retrieval measure of  precision@k, which considers the number of relevant matches in the top-k matchings \cite{manning2008introduction}. For each combination of the text similarity approaches and embeddings, we show precision@k  for different k's in Figure~\ref{fig:summarization_ranking}. We observe that our method with science embedding achieves the best performance compared to the baselines, once again showing not only the benefits of our method but also that of  incorporating domain knowledge.

\begin{figure}[t]
\centering
\includegraphics[width=0.48\textwidth]{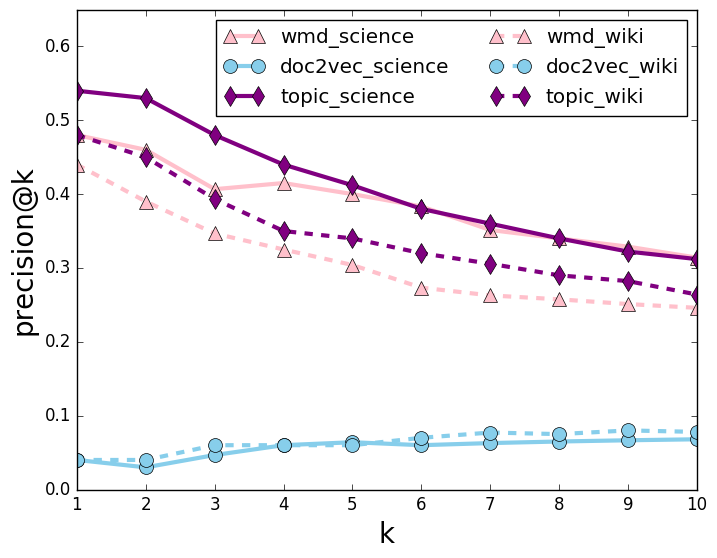}
\vspace{-5mm}
\caption{Summary-Article Matching}
\label{fig:summarization_ranking}
\end{figure}

\section{Discussion}
\noindent\textbf{Analysis of Results}. From the results of the two tasks we observe that our method outperforms two strong baselines. 
The reason for  WMD's poor performance  could be that the many uninformative words (those unrelated to the central topic)  make WMD overestimate the distance between the document-summary pair. As for doc2vec, its single vector representation may not be able to capture all the key topics of a document. A project could contain multifaceted information, e.g., a project to study how climate change affects grain production is related to both environmental science and agricultural science. 

\noindent\textbf{Effect of Topic Number}. The number of hidden topics $K$ is a hyperparameter in our setting. We empirically evaluate the effect of topic number in the task of concept-project mapping. Figure~\ref{fig:topic_num_analysis} shows the F1 scores and the standard deviations at different $K$.
\begin{figure}[t]
\centering
\includegraphics[width=0.48\textwidth]{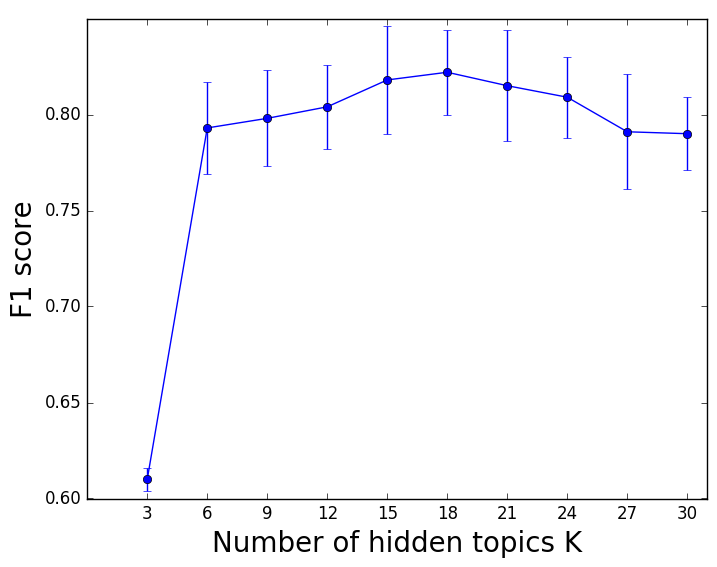}
\vspace{-4mm}
\caption{F1 score on concept-project matching with different topic numbers $K$}
\vspace{-5mm}
\label{fig:topic_num_analysis}
\end{figure}
As we can see, optimal $K$ is 18. When $K$ is too small, hidden topics are too few to capture key information in projects. Thus we can see that the increase of topic number from $3$ to $6$ brings a big improvement to the performance. 
Topic numbers larger than the optimal value degrade the performance since more topics incorporate noisy information.
We note that the performance changes are mild when the number of topics are in the range of [18, 31]. Since topics are weighted by their importance, the effect of noisy information from extra hidden topics is mitigated. 

\noindent\textbf{Interpretation of Hidden Topics}. We consider the summary-paper matching as an example with around 10 papers per category. We  extracted the hidden topics from each paper, reconstructed words with these topics as shown in Eq.~(\ref{eq:reconstruct}), and  selected the words which had the smallest reconstruction errors. These words are thus closely related to the hidden topics, and we call them \emph{topic words} to serve as an interpretation of the hidden topics. We visualize the cloud of such topic words on the set of papers about word sense disambiguation as shown in Figure~\ref{fig:wordcloud}. We see that the words selected based on the hidden topics cover key ideas such as \emph{disambiguation}, \emph{represent}, \emph{classification} and \emph{sentence}. This qualitatively validates the representation power of hidden topics. More examples are available in the supplementary material. 

We interpret this to mean that proposed idea of multiple hidden topics   captures the key information of a document. 
%The strong performance of our  approach has shown that its strength is the robustness of the measure of document-summary relevance. 
The extracted ``hidden topics'' represent the essence of documents, suggesting the appropriateness of our relevance metric  to measure the similarity between texts of different sizes. Even though our  focus in this study was the science domain we  point out that the results are more generally valid since we made no domain-specific assumptions.

\begin{figure}[t]
\centering
\includegraphics[width=0.45\textwidth]{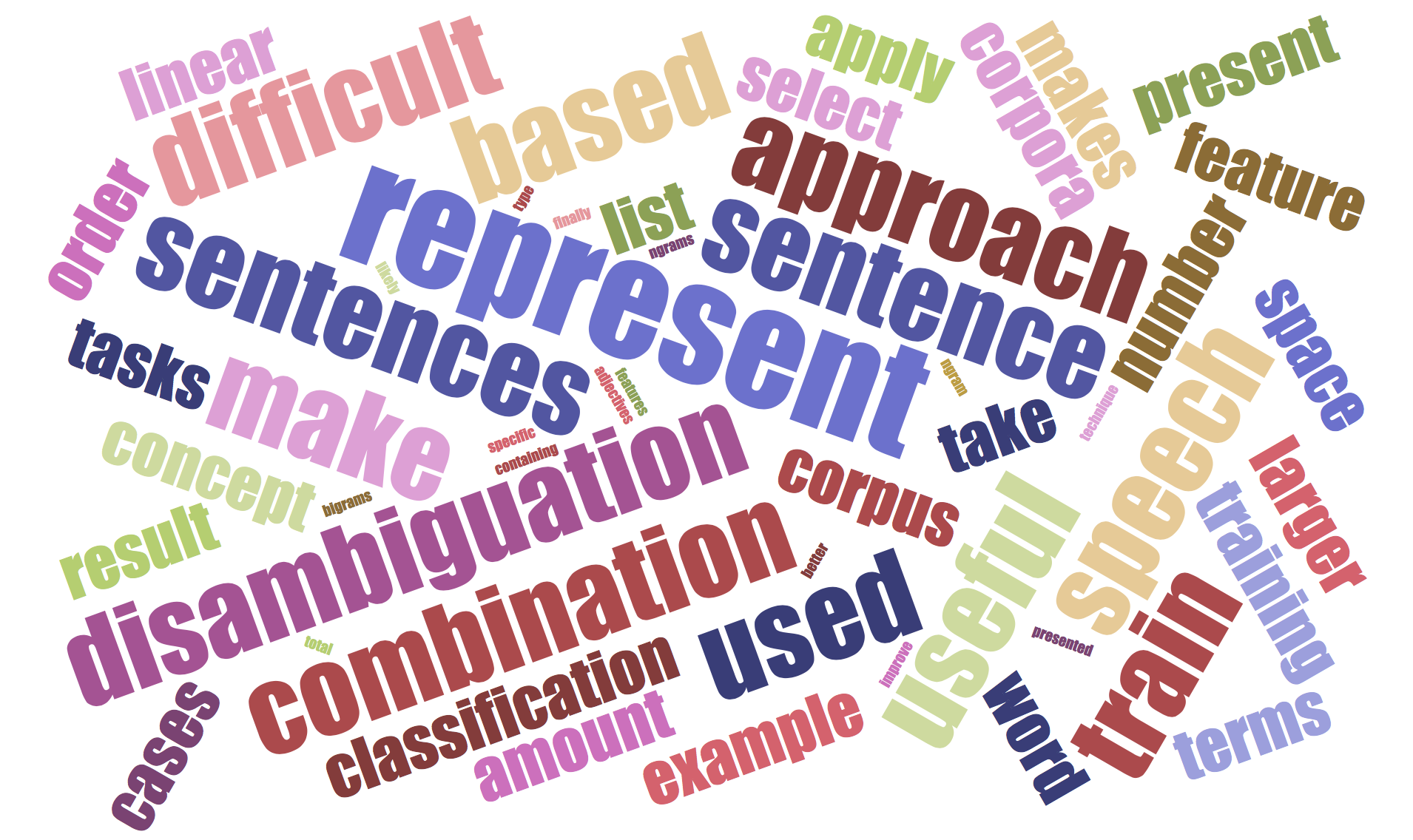}
\caption{Topic words from papers on word sense disambiguation}
\label{fig:wordcloud}
\end{figure}

\noindent\textbf{Varying Sensitivity to Domain}. As shown in  the results, the science-domain embeddings improved the classification of concept-project matching for the topic-based method  by $2\%$ in F1-score, WMD  by $8\%$ and doc2vec by $1\%$, thus underscoring the importance of domain-specific word embeddings.%We hypothesize that the effect of science-domain embeddings is more pronounced when using WMD. WMD, as a word embedding based method, is more sensitive to domain knowledge compared with other methods. 

Doc2vec is less sensitive to the domain, because it provides document-level representation. Even if some words cannot be disambiguated due to the lack of domain knowledge, other words in the same document can provide complementary information so that the document embedding does not deviate too much from its true meaning. 

Our method, also a word embedding method, is not as sensitive to domain as WMD. It is robust to the polysemous words with domain-sensitive semantics, since hidden topics are extracted in the document level. Broader contexts beyond just words provide complementary information for word sense disambiguation.

\section{Conclusion}
\label{sec:conclusion}
We propose a novel approach to matching documents and summaries. The challenge we address is to bridge the gap between detailed long texts and its abstraction with hidden topics. We incorporate domain knowledge into the matching system to gain further performance improvement. Our approach has beaten two strong baselines in two downstream applications, concept-project matching and summary-research paper matching.

\bibliography{acl2018}
\bibliographystyle{acl_natbib}

\newpage
\appendix
\onecolumn
\begin{center}
	\Large Supplementary Material
\end{center}

\section{Optimization Problem in Finding Hidden Topics}
We first show that problem (\ref{eq:2}) is equivalent to the optimization problem (\ref{eq:matrix}). The reconstruction of word ${\bf w}_{i}$ is $\tilde{{\bf w}}_{i}$, and $\tilde{{\bf w}}_{i}={\bf W}\tilde{{\bm\alpha}}_{i}$ where
\begin{align}
\label{app:reconstruct}
\tilde{{\bm\alpha}}_{i} = \argmin\limits_{{\bm\alpha}_{i}\in\mathbb{R}^{K}}\lVert{\bf w}_{i} - {\bf H}{\bm\alpha}_{i}\rVert_{2}^{2}.
\end{align}
Problem (\ref{app:reconstruct}) is a standard quadratic optimization problem which is solved by $\tilde{{\bm\alpha}}_{i}={\bf H}^{\dagger}{\bf w}_{i}$, where ${\bf H}^{\dagger}$ is the pseudoinverse of ${\bf H}$. With the orthonormal constraints on ${\bf H}$, we have ${\bf H}^{\dagger}={\bf H}^{T}$. Therefore, $\tilde{{\bm\alpha}}_{i}={\bf H}^{T}{\bf w}_{i}$, and $\tilde{{\bf w}}_{i}= {\bf H}{\bm\alpha}_{i} = {\bf H}{\bf H}^{T}{\bf w}_{i}$.

Given the topic vectors ${\bf H}$, the reconstruction error $E$ is defined as:
\begin{align}
\label{app:error}
\nonumber
E({\bf H}) &= \sum\limits_{i=1}^{n}\min\limits_{{\bm \alpha}_{i}}\lVert {\bf w}_{i}-{\bf H}{\bm\alpha}_{i}\rVert_{2}^{2} \\
\nonumber
&= \sum\limits_{i=1}^{n}\lVert {\bf w}_{i}-{\bf H}{\bf H}^{T}{\bf w}_{i}\rVert_{2}^{2} \\
&= \lVert {\bf W} - {\bf H}{\bf H}^{T}{\bf W} \rVert^{2}_{2},
\end{align}
where ${\bf W}=[{\bf w}_{1}, \ldots, {\bf w}_{n}]$ is a matrix stacked by word vectors $\{{\bf w}_{i}\}_{i=1}^{n}$ in a document. Now the equivalence has been shown between problem (\ref{eq:2}) and (\ref{eq:matrix}).

Next we show how to derive hidden topic vectors from the optimization problem (\ref{eq:matrix}) via Singular Value Decomposition. The optimization problem  is:
\begin{align*}
&\min\limits_{{\bf H}}\quad \lVert {\bf W}-{\bf H}{\bf H}^{T}{\bf W}\rVert^{2} \\
&\text{s.t.}\quad {\bf H}^{T}{\bf H} = {\bf I}
\end{align*}
Let ${\bf H}{\bf H}^{T}={\bf P}$. Then we have:
\begin{align*}
\sum\limits_{i=1}^{n}\lVert {\bf w}_{i}-{\bf P}{\bf w}_{i}\rVert^{2} &= \sum\limits_{i=1}^{n}({\bf w}_{i}-{\bf P}{\bf w}_{i})^{T}({\bf w}_{i}-{\bf P}{\bf w}_{i}) \\
&= \sum\limits_{i=1}^{n}({\bf w}_{i}^{T}{\bf w}_{i}-2{\bf w}_{i}^{T}{\bf P}{\bf w}_{i}+{\bf w}_{i}^{T}{\bf P}^{T}{\bf P}{\bf w}_{i}).
\end{align*}
Since ${\bf P}^{T}{\bf P}={\bf H}{\bf H}^{T}{\bf H}{\bf H}^{T}={\bf P}$, we only need to minimize:
$$\sum\limits_{i=1}^{n}(-2{\bf w}_{i}^{T}{\bf P}{\bf w}_{i}+{\bf w}_{i}^{T}{\bf P}{\bf w}_{i})=\sum\limits_{i=1}^{n}(-{\bf w}_{i}^{T}{\bf P}{\bf w}_{i}).$$
It is equivalent to the maximization of $\sum\limits_{i=1}^{n}{\bf w}_{i}^{T}{\bf P}{\bf w}_{i}$. 

Let $\text{tr}({\bf X})$ be the trace of a matrix ${\bf X}$, we can see that 
\begin{align}
\sum\limits_{i=1}^{n}{\bf w}_{i}^{T}{\bf P}{\bf w}_{i}&=\text{tr}({\bf W}^{T}{\bf P}{\bf W}) = \text{tr}({\bf W}^{T}{\bf H}{\bf H}^{T}{\bf W}) \\
\label{app:trace}
&= \text{tr}({\bf H}^{T}{\bf W}{\bf W}^{T}{\bf H}) \\
&= \sum\limits_{k=1}^{K}{\bf h}_{k}^{T}{\bf W}{\bf W}^{T}{\bf h}_{k}
\end{align}
Eq. (\ref{app:trace}) is based on one property of trace: $\text{tr}({\bf X}^{T}{\bf Y})=\text{tr}({\bf X}{\bf Y}^{T})$ for two matrices ${\bf X}$ and ${\bf Y}$.

The optimization problem (\ref{eq:matrix}) now can be rewritten as:
\begin{align}
\nonumber
&\max\limits_{\{{\bf h}_{k}\}_{k=1}^{K}}\qquad\sum\limits_{k=1}^{K}{\bf h}_{k}^{T}{\bf W}{\bf W}^{T}{\bf h}_{k} \\
\label{app:lagrangian}
&\text{s.t.}\qquad {\bf h}_{i}^{T}{\bf h}_{j}=1_{(i=j)}, \forall i,j
\end{align}

We apply Lagrangian multiplier method to solve the optimization problem (\ref{app:lagrangian}). The Lagrangian function $L$ with multipliers $\{\lambda_{k}\}_{k=1}^{K}$ is:
\begin{align*}
L &= \sum\limits_{k=1}^{K}{\bf h}_{k}^{T}{\bf W}{\bf W}^{T}{\bf h}_{k}-
\sum\limits_{k=1}^{K}(\lambda_{k}{\bf h}_{k}^{T}{\bf h}_{k}-\lambda_{k}) \\
&= \sum\limits_{k=1}^{K}{\bf h}_{k}^{T}({\bf W}{\bf W}^{T}-\lambda_{k}{\bf I}){\bf h}_{k} + \sum\limits_{k=1}^{K}{\lambda_{k}}
\end{align*}
By taking derivative of $L$ with respect to ${\bf h}_{k}$, we can get
\begin{align*}
\frac{\partial L}{\partial {\bf h}_{k}}=2({\bf W}{\bf W}^{T}-\lambda_{k}{\bf I}){\bf h}_{k}=0.
\end{align*}
If ${\bf h}_{k}^{*}$ is the solution to the equation above, we have 
\begin{align}
\label{app:eig}
{\bf W}{\bf W}^{T}{\bf h}_{k}^{*}=\lambda_{k}{\bf h}_{k}^{*},
\end{align}
which indicates that the optimal topic vector ${\bf h}_{k}^{*}$ is the set of eigenvectors of ${\bf W}{\bf W}^{T}$.

The eigenvector of ${\bf W}{\bf W}^{T}$ can be computed using Singular Value Decomposition (SVD). SVD decomposes matrix ${\bf W}$ can be decomposed as ${\bf W}={\bf U}{\bm\Sigma}{\bf V}^{T}$, where ${\bf U}^{T}{\bf U}={\bf I}$, ${\bf V}^{T}{\bf V}={\bf I}$, and ${\bm \Sigma}$ is a diagonal matrix. Because 
\begin{align*}
{\bf W}{\bf W}^{T}{\bf U}={\bf U}{\bm\Sigma\Sigma}^{T}{\bf U}^{T}{\bf U}={\bf U}{\bm\Sigma}',
\end{align*}
where ${\bm\Sigma}'={\bm\Sigma\bm\Sigma}^{T}$, and it is also a diagonal matrix. As is seen, ${\bf U}$ gives eigenvectors of ${\bf W}{\bf W}^{T}$, and the corresponding eigenvalues are the diagonal elements in ${\bm\Sigma}'$.

We note that not all topics are equally important, and the topic which recover word vectors $W$ with smaller error are more important. When $K=1$, we can find the most important topic which minimizes the reconstruction error $E$ among all vectors. Equivalently, the optimization in (\ref{app:lagrangian}) can be written as:
\begin{align}
\label{app:import}
{\bf h}_{1}^{*} = \argmax\limits_{{\bf h}_{1}:\lVert{\bf h}_{1}\rVert=1} {\bf h}_{1}^{T}{\bf W}{\bf W}^{T}{\bf h}_{1} = \argmax\limits_{{\bf h}_{1}:\lVert{\bf h}_{1}\rVert=1} \lambda_{1}{\bf h}_{1}^{T}{\bf h}_{1} = \argmax\limits_{{\bf h}_{1}}\lambda_{1}
\end{align}

The formula (\ref{app:import}) indicates that the most important topic vector is the eigenvector corresponds to the maximum eigenvalue. Similarly, we can find that the larger the eigenvalue $\lambda_{k}^{*}$ is, the smaller reconstruction error the topic ${\bf h}_{k}^{*}$ achieves, and the more important the topic is.

Also we can find that 
\begin{align*}
\lambda_{k}^{*}= {{\bf h}_{k}^{*}}^{T}{\bf W}{\bf W}^{T}{{\bf h}_{k}^{*}}=\lVert {{\bf h}_{k}^{*}}^{T}{\bf W}\rVert_{2}^{2}.
\end{align*}
As we can see, $\lVert {{\bf h}_{k}^{*}}^{T}{\bf W}\rVert_{2}^{2}$ can be used to quantify the importance of the topic $h_{k}$, and it is the unnormalized importance score $i_{k}$ we define in Eq. (\ref{eq:important}).

Henceforth, the $K$ vectors in $U$ corresponding to the largest eigenvalues are the solution to optimal hidden vectors $\{{\bf h}_{1}^{*},\ldots,{\bf h}_{K}^{*}\}$, and the topic importance is measured by $\{\lVert {{\bf h}_{1}^{*}}^{T}{\bf W}\rVert_{2}^{2}, \ldots, \lVert {{\bf h}_{K}^{*}}^{T}{\bf W}\rVert_{2}^{2}\}$.

\section{Interpretation of Hidden Topics}

\begin{figure}[htbp!]
\centering
\begin{minipage}[c]{0.48\textwidth}
\centerline{\includegraphics[width=\linewidth]{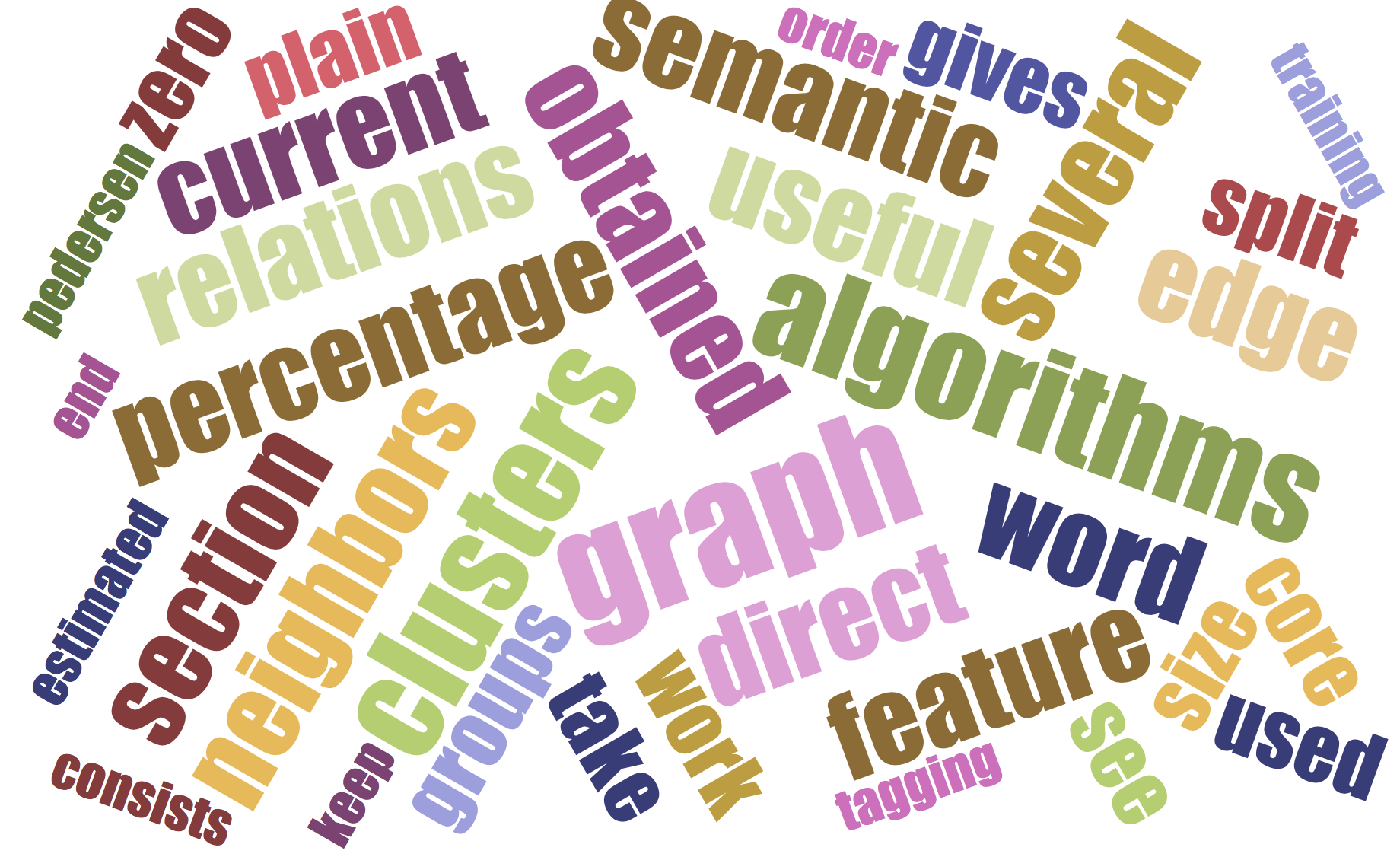}}
\centerline{\small{(a) Graph model to cluster senses}}
\label{fig:wordcloud_2}
\end{minipage}
\begin{minipage}[c]{0.48\textwidth}
\centerline{\includegraphics[width=\linewidth]{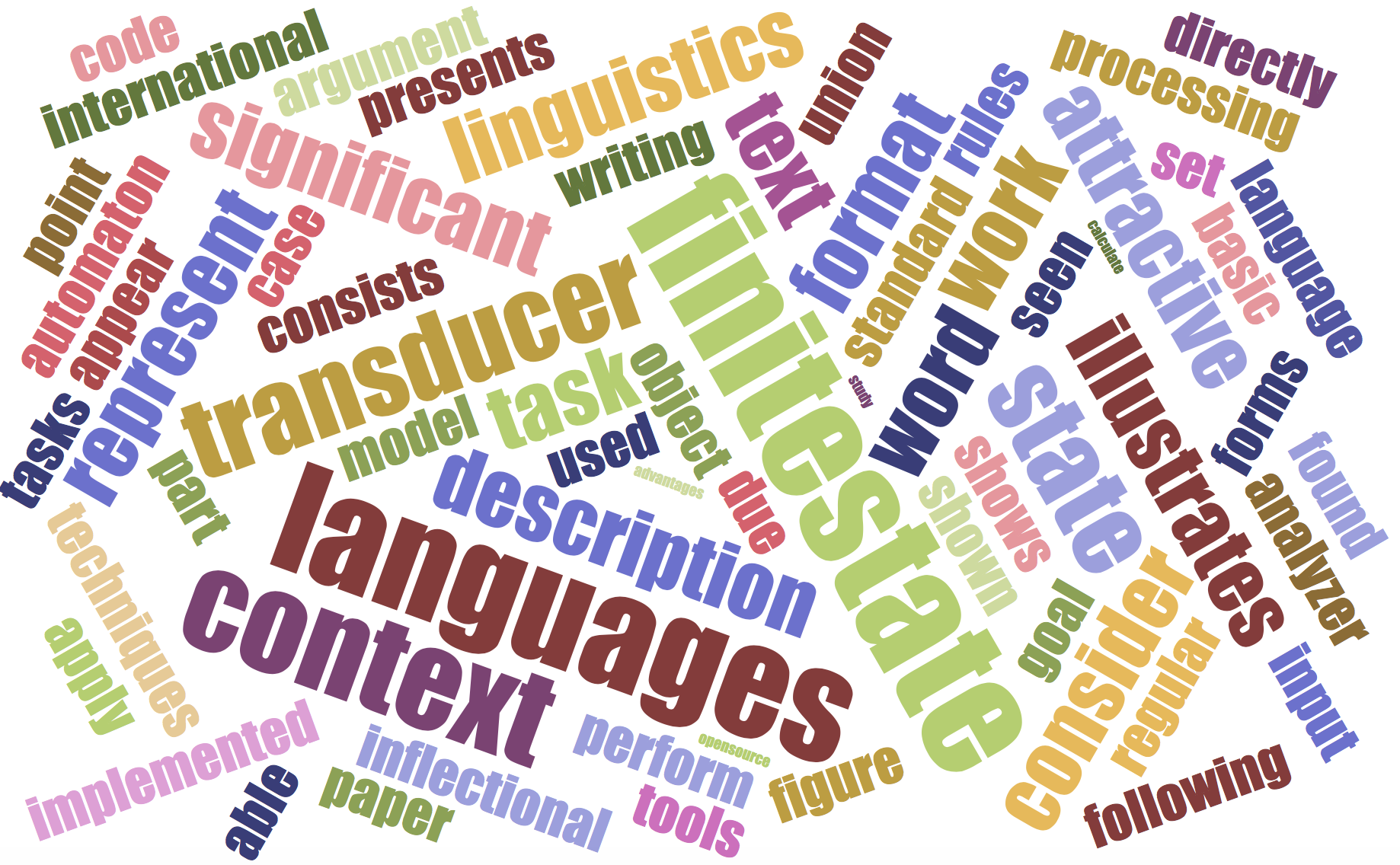}}
\centerline{\small{(b) Finite-state automaton as language analyzer}}
\label{fig:wordcloud_3}
\end{minipage}
\caption{Topic Word Visualization to Interpret Hidden Topics}
\label{fig:appendix_wordcloud}
\end{figure}

Mathematically the hidden topics are orthonormal vectors, and do not carry physical meaning. To gain a deeper insight of these hidden topics, we can establish the connections between topics and words. For a given paper, we can extract several hidden topics ${\bf H}^{*}$ by solving the optimization problem (\ref{eq:2}). 

For each word ${\bf w}_{i}$ in the document, we reconstruct $\tilde{{\bf w}}_{i}$ with hidden topic vectors $\{{\bf h}_{k}\}_{k=1}^{K}$ as below:
\begin{align*}
\tilde{{\bf w}}_{i} = {\bf H}{\bf H}^{T}{\bf w}_{i}
\end{align*}
The reconstruction error for word ${\bf w}_{i}$ is $\lVert{\bf w}_{i}-\tilde{{\bf w}}_{i}\rVert_{2}^{2}$. We select words with small reconstruction errors since they are closely relevant to extract topic vectors, and could well explain the hidden topics. We collect these highly relevant words from papers in the same category, which are natural interpretations of hidden topics. The cloud of these topic words are shown in Figure~\ref{fig:appendix_wordcloud}. The papers are about graph modeling to cluster word senses in Figure~\ref{fig:appendix_wordcloud}(a). As we can see, topic words such as \emph{graph}, \emph{clusters}, \emph{semantic}, \emph{algorithms} well capture the key ideas of those papers. Similarly, Figure~\ref{fig:appendix_wordcloud}(b) presents the word cloud for papers on finite-state automaton and language analyzer. Core concepts such as \emph{language}, \emph{context}, \emph{finite-state}, \emph{transducer} and \emph{linguistics} are well preserved by the extracted hidden topics.

\end{document}